\documentclass[runningheads]{llncs}
\usepackage{graphicx}
\usepackage{comment}
\usepackage{amsmath,amssymb} 
\usepackage{color}
\usepackage{multirow} 
\usepackage{array}
\newcolumntype{L}[1]{>{\raggedright\let\newline\\\arraybackslash}m{#1}}
\newcolumntype{C}[1]{>{\centering\let\newline\\\arraybackslash}m{#1}}
\newcolumntype{R}[1]{>{\raggedleft\let\newline\\\arraybackslash}m{#1}}
\usepackage{subfigure}
\usepackage{wrapfig}
\usepackage{hyperref}

\makeatletter
\DeclareRobustCommand\onedot{\futurelet\@let@token\@onedot}
\def\@onedot{\ifx\@let@token.\else.\null\fi\xspace}

\makeatother

\begin{document}
\pagestyle{headings}
\mainmatter
\def\ECCVSubNumber{----}  

\title{PatchAttack: A Black-box Texture-based Attack with Reinforcement Learning} 

\titlerunning{Black-box PatchAttack}
%
\author{Chenglin Yang \and
Adam Kortylewski \and
Cihang Xie \and
Yinzhi Cao \and
Alan Yuille
} 

\authorrunning{C. Yang et al.}
%
\institute{
Johns Hopkins University \\
\email{\{chenglin.yangw,cihangxie306,alan.l.yuille\}@gmail.com}\\
\email{\{akortyl1,yinzhi.cao\}@jhu.edu}
}
\maketitle

\begin{abstract}
Patch-based attacks introduce a perceptible but localized change to the input that induces misclassification.
A limitation of current patch-based black-box attacks is that they perform poorly for targeted attacks, and even for the less challenging non-targeted scenarios, they require a large number of queries.
Our proposed \textit{PatchAttack} is query efficient and can break models for both targeted and non-targeted attacks. 
PatchAttack induces misclassifications by superimposing small textured patches on the input image. 
We parametrize the appearance of these patches by a dictionary of class-specific textures. 
This texture dictionary is learned by clustering Gram matrices of feature activations from a VGG backbone.
\textit{PatchAttack} optimizes the position and texture parameters of each patch using reinforcement learning. 
Our experiments show that PatchAttack achieves $>99\%$ success rate on ImageNet for a wide range of architectures, while only manipulating $3\%$ of the image for non-targeted attacks and $10\%$ on average for targeted attacks.
Furthermore, we show that PatchAttack circumvents state-of-the-art adversarial defense methods successfully. The code is publicly available \href{https://github.com/Chenglin-Yang/PatchAttack}{\textcolor{cyan}{here}}.

\end{abstract}
\section{Introduction}
Computer vision models have achieved strong performance on image recognition tasks, however, they are known to be vulnerable against adversarial examples \cite{szegedy2013intriguing}.
Adversarial examples are modifications of images crafted to induce misclassification. 
Understanding the vulnerability of computer vision models to adversarial attacks has emerged as an important research area, providing opportunities for understanding and improving computer vision models.

Recent works have introduced very successful attacks in the white-box setting \cite{goodfellow2014explaining,madry2017towards,carlini17}, where both the network architecture and parameters are available to the attacker.
In real-world applications, a more common attacking scenario is that the attacker only has access to the model's input and the predicted output, \textit{e.g.} , attacking popular image analysis APIs \cite{googleapi,clarifai,hosseini2017deceiving,hosseini2017google,goodman2019cloud,goodman2020transferability} or self-driving cars \cite{bojarski2016end,pei2017deepxplore,tian2018deeptest,chernikova2019self,ranjan2019attacking,sitawarin2018darts,eykholt2018robust,Huang2020UPC}. 
This \textit{black-box} scenario is challenging because adversarial modification of the input must be computed without access to the loss gradient of the model.

Two paradigms have emerged for black-box attacks. \textit{Perturbation-based} methods introduce imperceptible changes to the image that are constrained to have a small norm but are typically applied to the whole input image \cite{papernot2016transferability,papernot2017practical,chen2017zoo,tu2019autozoom,bhagoji2018practical,squareattack}. 
Recently, several defense methods have shown that perturbation-based attacks can be successfully defended \cite{goodfellow2014explaining,kannan2018adversarial,madry2017towards,xie2019feature}.

In this paper, we study a complementary type of adversary, \textit{Patch-based black-box attacks}, introducing a perceptible (large norm) but localized change to the input.
In their pioneering work, Fawzi et al. \cite{fawzi2016measuring} show that superimposing monochrome black patches onto images generated by random search can successfully induce misclassificataions.
However, a major limitation of current patch-based black box attacks is that they perform poorly on targeted attacks and require large amounts of queries for non-targeted attacks (Experiments \ref{sec:exp-attack}). 

\begin{figure}[t]
    \centering
    \includegraphics[height=3cm]{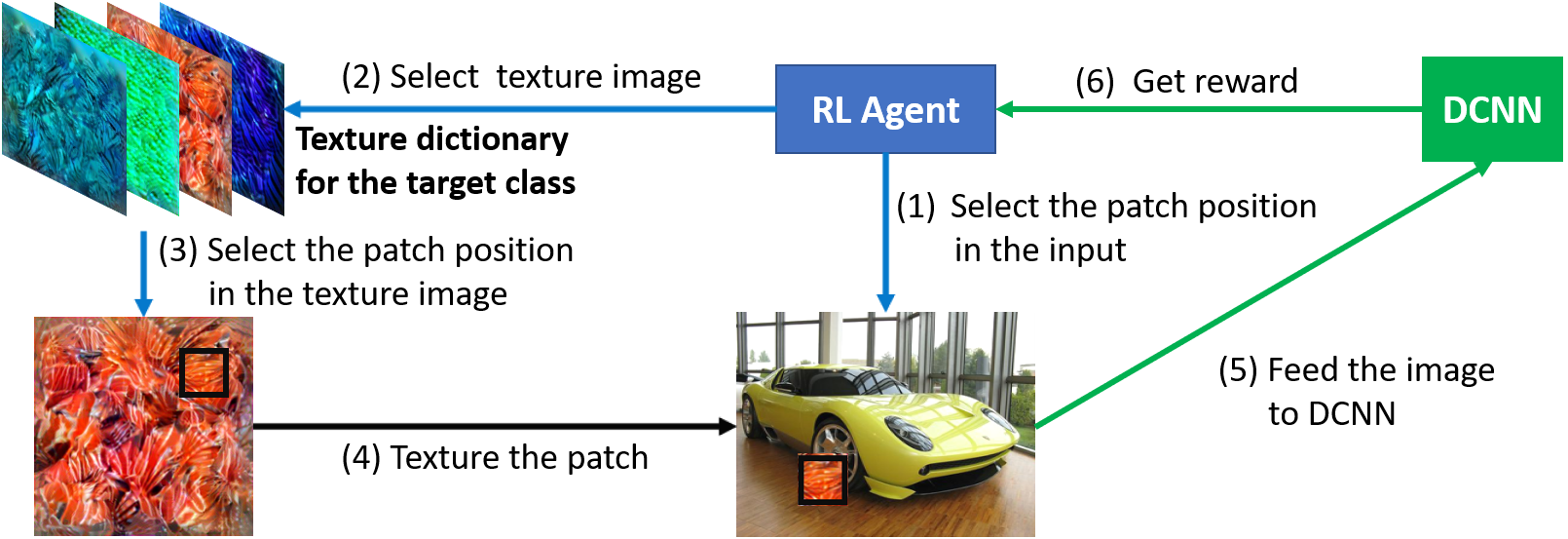}
    \caption{
    Illustration of our black-box texture-based patch attack via reinforcement learning (RL):
    (1) The RL agent selects the patch position in the input image and (2) selects a texture image for the target category (lionfish in this example) from our learned texture dictionary. (3) The agent selects a patch position in the texture image to extract a texture patch,  (4) which is then superimposed on the input image. (5) The adversarial image is fed into the deep convolutional network (DCNN). (6) The output scores of the DCNN is used to calculate the reward for the optimization of the agent. This six step process is repeated until the DCNN is attacked successfully.}
    \label{fig:my_label}
\end{figure}

In this work, we introduce \textit{PatchAttack}, a patch-based black-box attack that is query efficient and achieves very high  success  rates  for both  targeted  and  non-targeted  attacks. 
Our main contributions are two fold: 
1) We formulate the search over the position and shape of adversarial patches as a reinforcement learning problem. 
Hence, we define the attack as a decision-making process of an agent that interacts with its' environment (the model) by taking actions (placing patches in the image) and observing rewards (misclassification rates).
In this way, the parameter search is formulated as an optimization problem that is much more effective compared to random search strategies in terms of query efficiency.
2) Our experiments show that attacks with monochrome patches do not succeed as targeted attacks.
The intuition is that monochrome patches can remove information from the image, but do not add any information, which is critical to confuse a model in a targeted manner.
We overcome this limitation by introducing texture to the adversarial patches. To parameterize the texture efficiently, we learn a dictionary of class-specific textures by clustering Gram matrices of feature activations from a VGG backbone.
The texture dictionary enables a query-efficient search over the patch appearance and leads to very high success rates at targeted and non-targeted attacks, while also strongly reducing the image area that needs to be corrupted for a successful attack.

Our results on ImageNet \cite{russakovsky2015imagenet} show that PatchAttack achieves considerably higher success rates for targeted and non-targeted attacks compared to related work, while being more efficient in terms of the number of queries and the size of the attacked image area (on average only $3\%$ of the image needs to be modified for non-targeted and $<10\%$ for targeted attacks respectively).
Furthermore, we show that PatchAttack can successfully overcome Feature Denoising \cite{xie2019feature}, a state-of-the-art defense for perturbation based attacks.
Finally, we perform experiments with shape-based DCNNs \cite{geirhos2018imagenet} which were designed to overcome the texture bias of DCNNs trained on ImageNet, and hence should be more robust to PatchAttack.
Interestingly, we cannot observe any increased robustness of shape-based DCNNs, although PatchAttack is texture-based and the object shape is largely preserved in the adversarial images.
\section{Related Work}
Sparked by the seminal works of Szegedy et al. \cite{szegedy2013intriguing} adversarial machine learning has emerged as an important research area for understanding and improving deep neural networks.
In recent years, two complementary paradigms have emerged for black-box attacks, perturbation-based and patch-based attacks.
In this paper, we focus on patch-based black-box attacks, but we also provide a short review of perturbation-based black-box attacks as the search strategies for both attack types are related.

\textbf{Perturbation-based black box attacks.}
While we focus on black-box attacks with access to model output scores, attacks with even more limited access to model decisions only have been explored \cite{brendel2017decision,ilyas2018black,cheng2018query}. Such approaches often require lots of queries and are therefore difficult to apply in real-world applications.
Early work on perturbation-based black box attacks with access to prediction scores proposed to estimate model gradients with finite differences \cite{bhagoji2018practical,ilyas2018black,uesato2018adversarial,tu2019autozoom,ilyas2018prior,chen2017zoo}. 
In particular, these iterative attacks estimate gradients via sampling from a noise distribution around the feature point. 
While this approach is successful it requires large amounts of model queries.
Other approaches use evolutionary algorithms \cite{alzantot2019genattack,ilyas2018prior} or random search strategies \cite{guo2019simple,squareattack}, but still often require many queries to be successful.
A complementary approach is to compute transferable adversarial examples based on the gradient of substitute networks, 
\cite{papernot2016transferability,papernot2017practical,liu2016delving,dong2018boosting,xie2019improving,li2020learning,zhou2018transferable,shi2019curls,naseer2019cross}. 

The success of perturbation-based attacks has sparked an arms race between adversarial attacks and corresponding defense mechanisms \cite{goodfellow2014explaining,kannan2018adversarial,madry2017towards,xie2019feature}. 
A particularly successful defense method is feature denoising \cite{xie2019feature}, where the features in a neural network are denoised using non-local means during adversarial training. To the best of our knowledge, this defense mechanism has not been successfully broken yet. In our experiments, we show that our patch-based attack can defeat this defense successfully.

\textbf{Patch-based white box attacks} Tom et. al.~\cite{brown2017adversarial} proposed adversarial patch as a white-box attack to cause classification errors. They craft adversarial examples by superimposing a patch onto the input image. Given a deep network, the pattern of the patch is optimized using gradient descent. The trained patch performs well but overfits to the network architecture. As shown in their experiments, the patches trained from four different networks are still not able to confuse a fifth network with a high success rate when the patch area is less than $10\%$. We perform transferability experiments in Appendix E.

\textbf{Patch-based black box attacks.} 
The seminal work of Fawzi et al. \cite{fawzi2016measuring} introduced patch-based black box attacks. They don't optimize the pattern of the patches, and instead use the monochrome patches.
The position and shape of the rectangular patches was searched using Metropolis-Hastings sampling. We refer their attack as Hastings Patch Attack (HPA).
While their approach is successful, the random search strategy requires many queries. 
Furthermore, our experimental results show that using monochrome patches only to craft adversarial examples leads to very low success rates and even then requires to cover more than $70\%$ of the image(see Experiments \ref{sec:exp-attack}).

We introduce a patch-based black-box attack using textured patches that is optimized with reinforcement learning. Our approach is significantly more query efficient, achieves $>99\%$ success rates on targeted and non-targeted attacks and modifies only very small areas of the input image.
\section{Methods}

In this section, we first discuss the mathematical framework for patch-based adversarial attacks (Section \ref{methods_math}). 
In Section \ref{methods_MPA} we introduce our reinforcement learning (RL) framework for patch-based black-box attacks.
Finally, we discuss how the texture of adversarial patches can be optimized efficiently using RL by parametrizing the appearance of the patch with a class-specific texture dictionary learned by clustering Gram matrices of feature activations from a DCNN backbone (Section \ref{methods_TPA}).

\subsection{Mathematical Framework} 
\label{methods_math}

We denote a deep neural network as a function $ \mathbf{y} = \mathbf{f}(\mathbf{x}; \boldsymbol{\theta})$, where $\mathbf{x}$, $\boldsymbol{\theta}$ and $\mathbf{y}$ denote the input image, model parameters, and output score of the model after softmax.
To perform an adversarial attack we optimize an objective function:
\begin{equation} \label{eq-overview}
    \mathcal{L}\left(\mathbf{y}, y'\right), \text{\quad where\quad} \mathbf{y}= \mathbf{f}\left(\mathbf{g}\left(\mathbf{x}\right); \boldsymbol{\theta}\right),
\end{equation}
where $\mathcal{L}$ is the loss between the output of the neural network $\mathbf{y}$ and a target class $y'$ with $y$ denoting the ground truth label.
$\mathbf{g}(\mathbf{x})$ denotes the adversarial example obtained by perturbing $\mathbf{x}$.
For targeted attacks, the goal is to induce a high confidence score for the class $y'$ while non-targeted attacks, it is only to induce misclassifications.
In perturbation-based attacks, $\mathbf{g}(\cdot)$ modifies $\mathbf{x}$ at every pixel and the perturbation is constrained to have a small norm. 
In contrast, the only constraint for patch-based attacks is that the perturbation must be localized in a small region $\mathcal{E}$:
\begin{align} \label{eq-perturb}
    &\mathbf{g}\left(\mathbf{x}\right): \begin{cases} 
            x_{u, v} = \mathbf{T}\left(x_{u, v}\right), &\text{\quad if\quad} (u, v) \in \mathcal{E}\\
            x_{u, v} = x_{u, v}, &\text{\quad otherwise\quad} \\
    \end{cases} \\
    &\mathcal{E} = \mathbf{s}\left(\mathbf{x},\mathbf{f}(\cdot,\mathbf{\theta}), \mathcal{S}\right) 
    \;\subseteq\; \left\{\left(u, v\right) | u \in \left[0, H\right), v \in \left[0, W\right) \right\}
\end{align}
$H$, $W$ are the height and width of a image, $u, v$ are the pixel coordinates. $\mathbf{T}(\cdot)$ is the transformation function applied to pixels inside $\mathcal{E}$.
To determine $\mathcal{E}$, a search mechanism $\mathbf{s}(\cdot)$ is defined over a search space $\mathcal{S}$ of potential image areas. 
The optimal region $\mathcal{E}^*$ depends on the input image $\mathbf{x}$ and the neural network $\mathbf{f}(\cdot,\theta)$. HPA uses Metropolis Hastings sampling to search the space $\mathcal{S}$ defined in Eq~\ref{space_MPA}.

\subsection{Patch Search with Reinforcement Learning}
\label{methods_MPA}

In this section, we propose our Monochrome Patch Attack (MPA). In general, this black-box attack uses monochrome rectangular patches which do not have patterns but have variable sizes and positions. 
We formulate the search over the position and size of adversarial patches as a reinforcement learning problem. The environment consists of $\mathbf{x}$ and $\mathbf{f}(\cdot,\theta)$, and an agent $\mathbb{A}$ is trained to sequentially place monochrome patches in the input image. The search space is defined as:
\begin{align}
    \mathcal{S} = \{(u_{1}^{1},v_{1}^{2},u_{1}^{3},v_{1}^{4}, \cdots, u_{C}^{1},v_{C}^{2},u_{C}^{3},v_{C}^{4}) \}\label{space_MPA}
\end{align}
where $C$ is the number of patches and each element in this set represents the coordinates of $C$ pair of opposite corner points with each pair determining one rectangular region. $\mathcal{S}$ has $4C$ dimensions therefore we set the agent to take $4C$ actions in sequence to generate $\mathbf{a} \in \mathcal{S}$. We formulate the attack in the following:
\begin{align}
    &\mathbb{A}(\mathbf{\theta}_{\mathbb{A}}): P(a_{t}|(a_{1},\cdots,a_{t-1}), \mathbf{f}(\cdot;\mathbf{\theta}), \mathbf{x})  \label {agent}
    \qquad\qquad\,\,\, t = \{1, \cdots, 4C\}\\
    &\mathbf{r} = \begin{cases} \label{Gray-MPA_r}
        \ln y' - \mathbf{A}\left(\mathbf{a}\right) / \sigma^2, &\text{target attack} \\
        \ln (1-y) - \mathbf{A}\left(\mathbf{a}\right) / \sigma^2, &\text{non-target attack}
    \end{cases}\\
    &\text{MPA}: \begin{cases} \label{Gray-MPA_L}
        &\mathcal{E} = \mathbf{J}\left(\mathbf{a}\right)\\
        &\mathbf{T}\left(x_{u,v}\right) = 0 \\
        &\mathcal{L} = -\mathbf{r}\cdot \ln \mathbf{P}
    \end{cases}
\end{align}




\noindent Similar to ~\cite{ren2017deep,shu2019identifying}, we define $\mathbb{A}$ to be a combination of an LSTM and a fully connected layer which represents a policy network with $\mathbf{\theta}_{\mathbb{A}}$ being its parameters. At step $t$, the environment state is determined by previous actions, the deep network and the input. The agent outputs the probability distribution over the possible actions for step $t$ as shown in Eq~\ref{agent}. Then it samples one action and records the probability of the sampled action. In the end, this agent generates an action sequence $\mathbf{a}$ and the probability sequence $\mathbf{P}$ recording sampling these actions at each step.
$\mathbf{J}(\cdot): \mathbf{a} \rightarrow \mathcal{E}$ is the function transferring $\mathbf{a}$ to the areas formed by the $C$ patches. 
The values of pixels in $\mathcal{E}$ are changed to $0$ as shown in Eq~\ref{Gray-MPA_r}. Since $\mathbf{x}$ is normalized, the patch color is gray.
The reward $\mathbf{r}$ for the agent is defined in Eq~\ref{Gray-MPA_r}, where $\mathbf{A}(\cdot)$ calculates the area of $\mathcal{E}$ 
and $\sigma$ controls the penalty on this area. 
The loss function to optimize $\boldsymbol{\theta}_{\mathbb{A}}$ is shown in Eq~\ref{Gray-MPA_L}.

Based on this framework, we further extend the search space to 
\begin{align}
    \mathcal{S} = \{(u_{1}^{1},v_{1}^{2},u_{1}^{3},v_{1}^{4}, R_{1}^{5}, G_{1}^{6},B_{1}^{7}, \cdots, u_{C}^{1},v_{C}^{2},u_{C}^{3},v_{C}^{4},R_{C}^{5}, G_{C}^{6},B_{C}^{7} )\}
\end{align}
where $R,G,B$ represent the values of the RGB channels of the patches, splitting MPA into two variants MPA\_Gray and MPA\_RGB. 




\subsection{Texture-based Patch Attacks} \label{methods_TPA}

Monochrome Patch Attacks (MPAs) are powerful in non-targeted setting, however, in targeted setting their performance is not satisfying (Experiments \ref{sec:exp-attack}), because MPAs only remove information at some parts of the image. The lack of additional input signals prevents MPAs from performing targeted attacks. 
However, we observe that MPA\_RGB achieves superior performance compared to MPA\_Gray (Table \ref{Tab:Non-target attack}), motivating our texture-based patch attacks.

\begin{figure}[!ht]
    \centering
    \includegraphics[width=\textwidth]{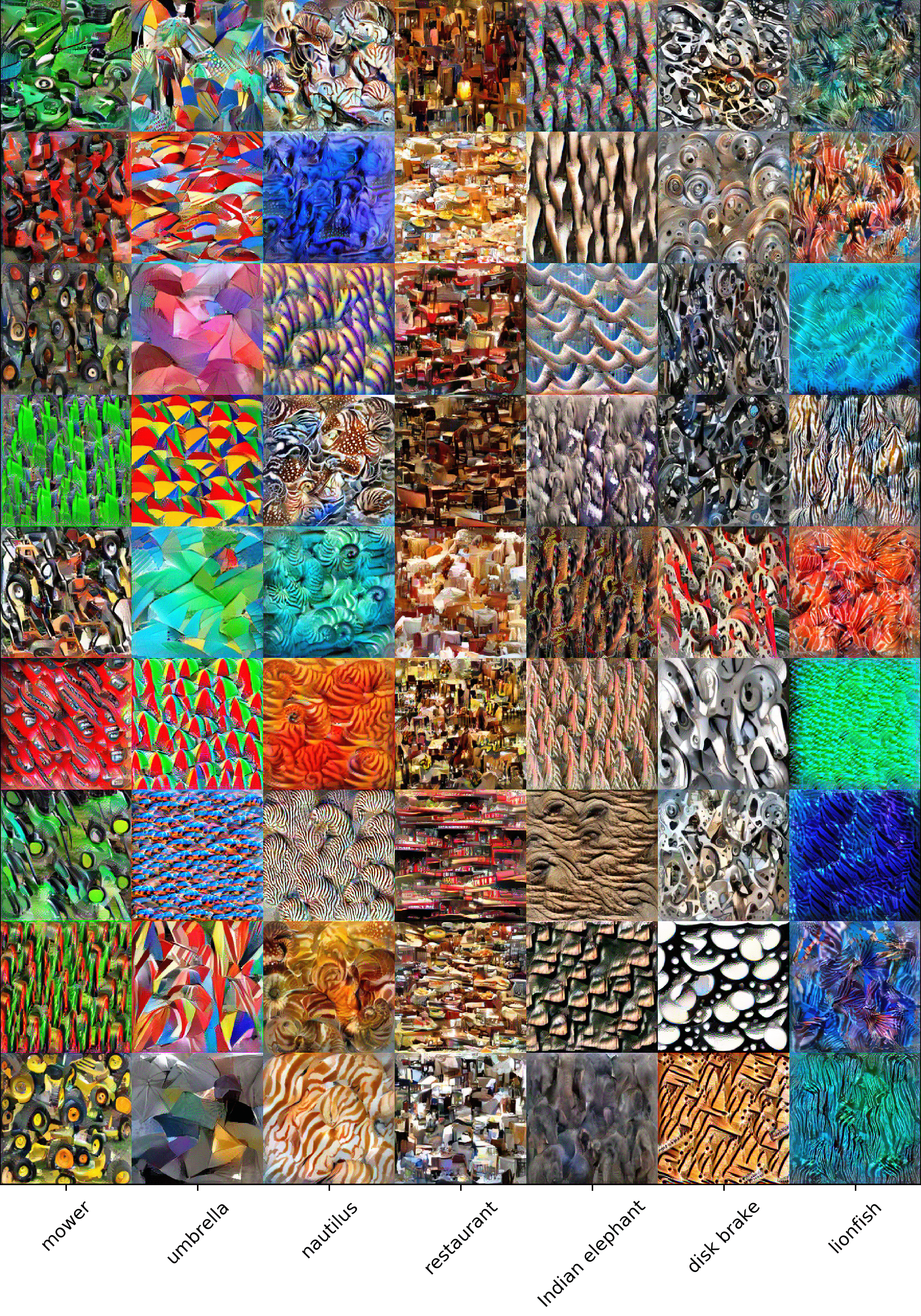}
    \caption{Examples in our designed texture dictionary.}
    \label{fig:texture dictionary}
\end{figure}

\textbf{A Class-specific Dictionary of Adversarial Textures.} 
A major challenge when adding texture to patches is to find an efficient parameterization of the texture to retain fast and query efficient attacks.
Our solution is to build a class-specific texture dictionary, where the patch patterns can be searched from.
Each dictionary element represents a prototypical adversarial texture of a target class.
Hence, to attack models trained on ImageNet, we build a dictionary with $1000$ different categories, corresponding to the $1000$ object classes in ImageNet. 
Each category has 30 different texture images whose contents are extracted from the ImageNet training set (see Figure \ref{fig:texture dictionary} for examples of dictionary elements).

We generate the texture dictionary using a four step process: First, we extract class-specific textures from a set of images of the target class. Inspired by style transfer~\cite{gatys2016image,gatys2015texture}, we use VGG19~\cite{simonyan2014very} as the backbone for extracting texture information from images. 
Let $\mathbb{D}$ be the fully convolutional part of VGG19 pre-trained on ImageNet and it consists of $5$ blocks. Let $\mathbf{F}_{i}^{j}$ be the feature maps from the $j$th convolution layer in the $i$th block of $\mathbb{D}$, and $\mathbf{G}_{i}^{j}$ be the corresponding Gram matrix of the feature activations. 
Following the approach of style transfer, we feed each image into $\mathbb{D}$ and compute the following gram matrices: $\mathbf{G}_{1}^{2}$, $\mathbf{G}_{2}^{2}$, $\mathbf{G}_{3}^{2}$ and $\mathbf{G}_{4}^{2}$.
Subsequently, we flatten all gram matrices and concatenate them into one vector $\mathbf{\bar{G}}$ that encodes the texture information. 

In a natural image, not all the regions are equally important for the final classification. 
Often the backgrounds or other objects are not of interest. 
Therefore, we need to extract only the texture of relevant objects in images and hence make the extracted $\mathbf{\bar{G}}$ more semantically meaningful and increase the transferability among different deep networks.
In order to locate the relevant information in each image, we perform Grad-CAM~\cite{selvaraju2017grad} on VGG19 for each image and mask out irrelevant regions of the image before texture extractions.

The third step is to generate the texture embedding. For each category, we conduct k-means algorithms on $\mathbf{\bar{G}}$s and use the $30$ calculated clusters as the texture embedding for that category $\left\{\mathbf{\bar{G}}_{c}^{1}, \ldots, \mathbf{\bar{G}}_{c}^{30}\right\}$, in order to increase the generalization property while maintaining the diversity of the embedding. The fourth part is to generate texture images from the $30\times 1000 \mathbf{\bar{G}}$s to build the dictionary. For each $\mathbf{\bar{G}}$, we optimize a texture image $\mathbf{t}$ starting from random noise according to the objective function $\mathcal{L}_{\text{texture}} = \lambda(\mathbf{\bar{G}} - \mathbf{G}_{\mathbf{t}})^{2}$, where $\mathbf{G}_{\mathbf{t}}$ and $\lambda$ denote the feature embedding of $\mathbf{t}$ and the weight. See details in Section \ref{texture dictionary setting}. 


\textbf{Integrating the Texture Dictionary into Patch Attack} 
Combining the generated texture dictionary, we propose Texture-based Patch Attack (TPA). Compared with MPA, the patches with the optimized locations in TPA are textured and provide more additional information, making TPA a powerful attack in both the non-targeted and targeted setting.

There are two updates from MPA to TPA. First, the search space is updated:
\begin{align}
    \mathcal{S} = \{(u_{1}^{1},v_{1}^{2},i_{1}^{3},u_{1}^{4}, v_{1}^{5}, \cdots, u_{C}^{1},v_{C}^{2},i_{C}^{3},u_{C}^{4},v_{C}^{5})\}
\end{align}
where $i_{c}^3$ indexes the texture image in category $y'$ used to texture the $c$th patch. $u_{c}^{1}$, $v_{c}^{2}$ determines the patch position in $\mathbf{x}$, represented by $\mathbf{J}(\cdot)_{t}^{1}$ in Eq.~\ref{Gray-TPA_L}. While $u_{c}^{4}$, $v_{c}^{5}$ denote the patch position in the $i_{c}^{3}$th texture image where we crop the patterns as the texture of the attacking patches, represented by $\mathbf{J}(\cdot)_{t}^{2}$. Note that in TPA, $C$ instead of $1$ agents are trained one after another. The $c$th agent's task is to find a position to put one more textured patch onto the image with $(c-1)$ patches already superimposed by the previous agents. The number of agents required to perform an successful attack $C$ varies for different $\mathbf{x}$. We set a maximum number of the patches allowed to place and stop training new agents if the attack is already successful or $C$ reaches the limit. The second update is that there is no penalty term on the patch area in Eq.~\ref{Gray-MPA_L}, since the size of each patch an agent can place and texture is pre-fixed. Different from MPA, the total area of the attacking regions is well controlled. 
\begin{align}
    &\text{TPA}: \begin{cases} \label{Gray-TPA_L}
        &\mathcal{E} = \mathbf{J}_{t}^{1}(u_{1}^{1},v_{1}^{2}, \cdots, u_{C}^{1},v_{C}^{2})\\
        &\mathbf{T}\left(x_{u,v}\right) = \mathbf{J}_{t}^{2}((i_{1}^{3},u_{1}^{4},v_{1}^{5}, \cdots, i_{C}^{3},u_{C}^{4},v_{C}^{5})) \\
    \end{cases}
\end{align}

\section{Experiments}
\label{sec:exp}


We conduct experiments on a challenging dataset, ILSVRC2012 \cite{russakovsky2015imagenet}, a popular subset of the ImageNet database \cite{deng2009imagenet}. It consists of $1.3$M training images and $50$k testing images with high resolution. There are $1000$ object categories in total, which are distributed approximately uniformly in the training set and strictly uniformly in the testing set. The networks against which we perform the attacks include ResNet~\cite{he2016deep}, DenseNet~\cite{huang2017densely}, ResNeXt~\cite{xie2017aggregated} and MobileNetV2~\cite{sandler2018mobilenetv2}. Since our texture dictionary are built through VGG~\cite{simonyan2014very} backbone, we do not involve this network to demonstrate the transferability of the texture images in the dictionary. For MPA and TPA, we conduct baseline subtraction on the rewards for the agents, and adopt early stopping when the difference of $\ln r$ averaged on $3$ consecutive iterations is less than $1\times 10^{-4}$, where $r$ is reward. 

\subsection{Texture Dictionary Setting} \label{texture dictionary setting}

\begin{figure}[t]
    \centering
    \includegraphics[width=\textwidth]{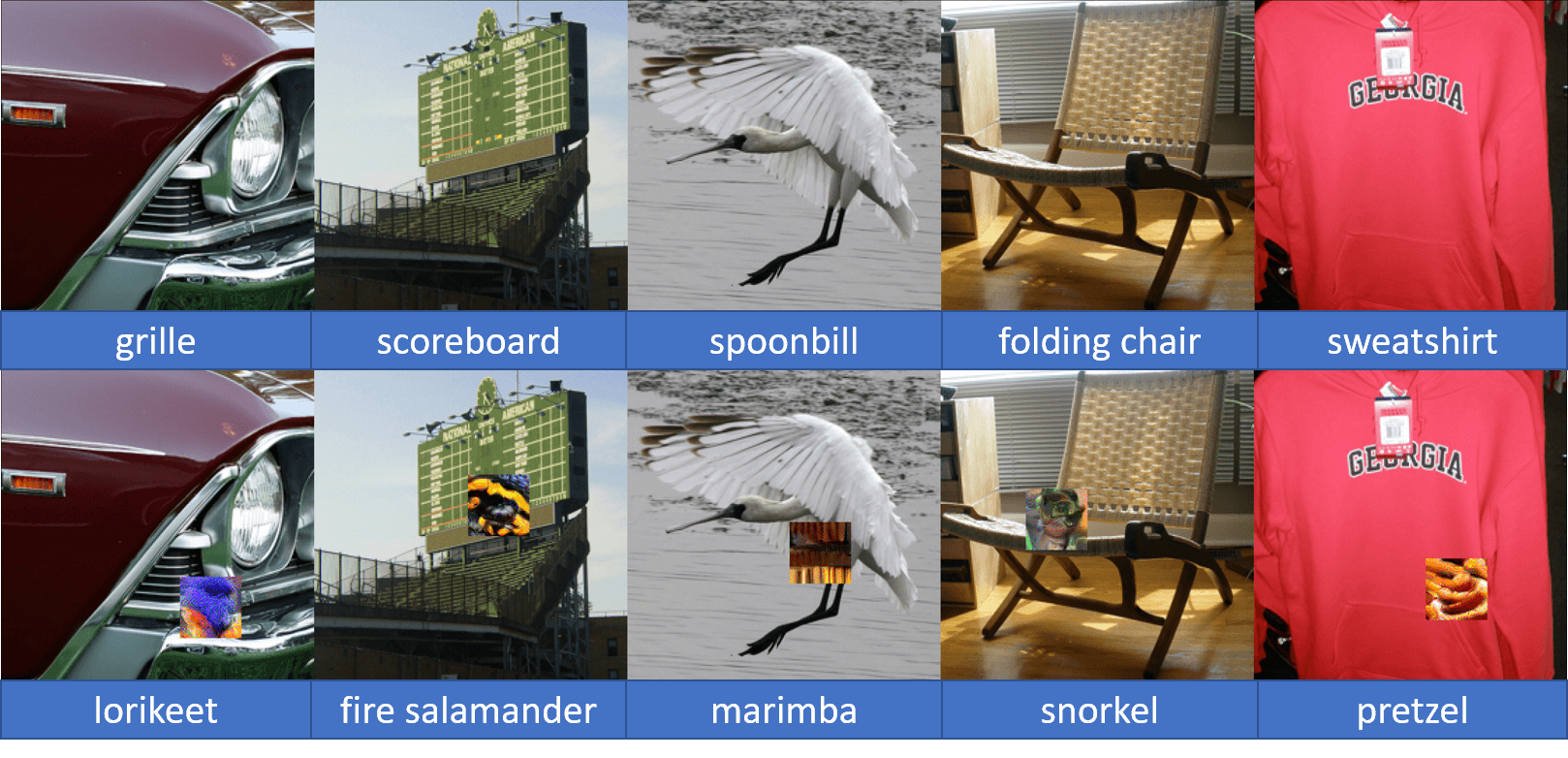}
    \caption{Adversarial examples generated by TPA\_N1\_4\%. The first blue row shows the ground truth labels while the second row the predictions of ResNet50. Each attacking patch is textured by the texture dictionary, taking $4\%$ of the overall area.}
    \label{fig:demo_attack}
\end{figure}

The texture dictionary is built upon the training set in ILSVRC2012. 
All max pooling layers in the extractor $\mathbb{D}$ are replaced by average pooling layers with the kernel and stride sizes both being $2$. 
The Grad-CAM is applied, using the feature map responses of the $5$th convolution block of VGG19 to generate a attention map whose values are then normalized between $0$ and $1$. 
We consider the region with attention scores larger than the threshold $0.8$ as useful. In the generation of texture images, the Adam optimizer is used with the starting learning rate $0.01$. The total iteration number is $10000$ and the weight $\lambda$ in $\mathcal{L}_{\text{texture}}$ is $1\times10^{6}$. The texture dictionary is constructed with a two-level index structure 
with first-level key indexing the categories and sub-level indexing $30$ texture images of a category. 
Figure \ref{fig:texture dictionary} and Figure \ref{fig:demo_attack} shows some example texture images and how the texture dictionary is utilized to perform attacks. 


\subsection{Attack Performance}
\label{sec:exp-attack}
In this section, we demonstrate the effectiveness of our proposed attacks, Mono-chrome Patch Attack (MPA) and Texture-based Patch Attack (TPA). The baseline is Hastings Patch Attack (HPA)~\cite{fawzi2016measuring}.
Both HPA and MPA\_Gray superimpose three gray rectangular patches onto the image. Each patch can take an arbitrary aspect ratio and an arbitrary scale. The updates of MPA\_RGB is that the patch color is optimized. 
In TPA, we adopt the square patches and fix their sizes. We set a maximum number of the patches which the algorithm can superimpose. Therefore, the actual number for each image varies. For example, TPA\_N$4$\_$4\%$ indicates that each patch occupies $4\%$ of the image and the maximum patch number is $4$. This means TPA is able to control the maximum allowed area of an image to be occluded by the patches. There is no such limit on HPA and MPA, which means TPA attacks are better controllable. Additionally, we provide comparisons between Hastings sampling and reinforcement learning in Appendix D.

\newcommand{\colwidthA}{2.5cm}
\newcommand{\colwidthB}{2.1cm}
\newcommand{\colwidthC}{1.5cm}
\begin{table*}[!t]
\tiny
\centering{
\caption{
Experimental results of the non-targeted attacks on a $1000$ images randomly selected from the ILSVRC2012 validation set. The maximum allowed query number is $10000$. Acc., Avg\_area, Avg\_qry denotes the classification accuracy, average area percentage occluded by the patches, average queries, respectively}
\small
\setlength{\tabcolsep}{0.08cm}
\begin{tabular}{|l||C{\colwidthA}|C{\colwidthB}|C{\colwidthB}|C{\colwidthB}|}
\hline
Network & Attack & Acc. (\%) & Avg\_area (\%) & Avg\_qry \\
\hline\hline

\multirow{6}{*}{ResNet50} & $-$ & $72.80$ & $-$ & $-$  \\
\cline{2-5}
& {\bf HPA} & $0.40$ & $18.05$ & $10000$  \\
\cline{2-5}
{} & {\bf MPA}\_Gray & $0.00$ & $6.57$ & $9659$ \\
\cline{2-5}
{} & {\bf MPA}\_RGB & $0.00$ & $5.41$ & $9681$   \\
\cline{2-5}
{} & {\bf TPA}\_N$4$\_$4\%$ & $0.30$ & $5.06$ & $1137$ \\
\cline{2-5}
{} & {\bf TPA}\_N$8$\_$2\%$ & $0.30$ & $3.10$ & $983$ \\

\hline\hline

\multirow{6}{*}{DenseNet121} & $-$ & $74.10$ & $-$ & $-$  \\
\cline{2-5}
& {\bf HPA} & $0.10$ & $19.82$ & $10000$  \\
\cline{2-5}
{} & {\bf MPA}\_Gray & $0.00$ & $6.87$ & $9624$ \\
\cline{2-5}
{} & {\bf MPA}\_RGB & $0.00$ & $5.73$ & $9696$   \\
\cline{2-5}
{} & {\bf TPA}\_N$4$\_$4\%$ & $0.50$ & $5.13$ & $1195$ \\
\cline{2-5}
{} & {\bf TPA}\_N$8$\_$2\%$ & $0.30$ & $3.13$ & $1001$ \\

\hline\hline

\multirow{6}{*}{ResNeXt50} & $-$ & $76.20$ & $-$ & $-$  \\
\cline{2-5}
& {\bf HPA} & $0.80$ & $19.22$ & $10000$  \\
\cline{2-5}
{} & {\bf MPA}\_Gray & $0.00$ & $7.88$ & $9748$ \\
\cline{2-5}
{} & {\bf MPA}\_RGB & $0.00$ & $6.23$ & $9752$   \\
\cline{2-5}
{} & {\bf TPA}\_N$4$\_$4\%$ & $0.70$ & $5.21$ & $1280$ \\
\cline{2-5}
{} & {\bf TPA}\_N$8$\_$2\%$ & $0.50$ & $3.25$ & $1088$ \\

\hline\hline

\multirow{6}{*}{MobileNet-V2} & $-$ & $68.80$ & $-$ & $-$  \\
\cline{2-5}
& {\bf HPA} & $0.20$ & $16.61$ & $10000$  \\
\cline{2-5}
{} & {\bf MPA}\_Gray & $0.00$ & $5.35$ & $9578$ \\
\cline{2-5}
{} & {\bf MPA\_RGB} & $0.00$ & $4.11$ & $9603$   \\
\cline{2-5}
{} & {\bf TPA}\_N$4$\_$4\%$ & $0.30$ & $4.63$ & $862$ \\
\cline{2-5}
{} & {\bf TPA}\_N$8$\_$2\%$ & $0.30$ & $2.74$ & $756$ \\

\hline

\end{tabular}}
\label{Tab:Non-target attack}
\end{table*}

The experimental results in non-targeted setting are summarized in Table \ref{Tab:Non-target attack}. In terms of accuracy drops, all the attacks achieve good performances against all the networks, decreasing their classification accuracy down to less than $1\%$. However, in terms of the average attacked area, HPA occludes $18.05\%$, $19.82\%$, $19.22\%$ and $16.61\%$ of the original images against the $4$ architectures, while our MPA\_RGB only occludes $5.41\%$, $5.73\%$, $6.23\%$ and $4.11\%$ respectively. Comparing MPA\_RGB with MPA\_Gray, we find that optimizing the RGB channel of the patches in MPA decreases the occluded area averaged over all the cases, from $6.67\%$ to $5.37\%$. This proves that increasing the optimization dimensions and improving the complexity of the patches is beneficial, motivating us to texture these patches. TPA occludes the least area of the image with $3.10\%$, $3.13\%$, $3.25\%$ and $2.74\%$ against the different networks. TPA\_N$8$\_$8\%$ works better than TPA\_N$4$\_$4\%$. 
In terms of query numbers, HPA is inefficient and always uses the whole query budget since it takes $10000$ samples and chooses the best one. From MPA to TPA, the algorithm becomes more and more efficient with the query times dropping from $9652$ to $957$.


For the more challenging targeted setting 
the experimental results are reported in Table \ref{Tab:Target attack}. Before performing the attacks, all the networks have a target accuracy not larger than $0.1\%$. It is observed that although HPA increases target accuracy to $23.05\%$ on average, it occluded $71.54\%$, $71.68\%$, $72.57\%$ and $69.45\%$ of the image in the four attacking cases, failing to be considered a successful attack algorithm. For MPA, we use the RGB version since it has been proved to be superior than the gray version in Table \ref{Tab:Non-target attack}. Although the MPA can only increase the target accuracy to $26.53\%$, it occludes much less areas than HPA with an average proportion $17.08\%$. 
On the contrary, our TPA achieves high performances. TPA\_N$10$\_$4\%$ is able to increase the target accuracy to $99.70\%$, $99.90\%$, $99.70\%$ and $99.90\%$ against the different architectures. The other $2$ variant TPA\_N$10$\_$2\%$ and TPA\_N$10$\_$10\%$ corresponds to two different requirements for the attack. The first one provides the smaller occlusion area as it uses $7.80\%$, $7.87\%$, $7.59\%$ and $7.78\%$ of the areas respectively to increase the target accuracy to $97.70\%$ on average. The second one is more query-efficient as it takes $3747$, $3970$, $3538$ and $4422$ queries and obtain an average target accuracy $100\%$. 

In both the non-targeted and non-targeted settings, MPA and TPA are superior to HPA to a large margin. TPA is the best attack among all the perspectives including the accuracy/target\_accuracy, occluded areas and query efficiency.

\renewcommand{\colwidthA}{2.5cm}
\renewcommand{\colwidthB}{2.1cm}
\renewcommand{\colwidthC}{1.5cm}
\begin{table*}[!ht]
\tiny
\caption{
Experimental results of the targeted attacks on a $1000$ images randomly selected from the ILSVRC2012 validation set. The maximum allowed query number is $50000$. The target label for each image is difference from its ground truth label. T\_acc., Avg\_area, Avg\_qry denotes the classification accuracy on target labels, average area percentage occluded by the patches, average queries, respectively}
\centering{
\small
\setlength{\tabcolsep}{0.08cm}
\begin{tabular}{|l||C{\colwidthA}|C{\colwidthB}|C{\colwidthB}|C{\colwidthB}|C{\colwidthB}|}
\hline
Network & Attack & T\_acc. (\%) & Avg\_area (\%)  & Avg\_qry \\
\hline\hline

\multirow{6}{*}{ResNet50} & $-$ & $0.10$ & $-$ & $-$  \\
\cline{2-5}
& {\bf HPA} & $23.20$ & $71.54$ & $50000$  \\
\cline{2-5}
{} & {\bf MPA}\_RGB & $25.90$ & $18.45$ & $28361$   \\
\cline{2-5}
{} & {\bf TPA}\_N$10$\_$2\%$ & $97.60$ & $7.80$ & $15728$ \\
\cline{2-5}
{} & {\bf TPA}\_N$10$\_$4\%$ & $99.70$ & $9.97$ & $8643$ \\
\cline{2-5}
{} & {\bf TPA}\_N$10$\_$10\%$ & $100.00$ & $15.36$ & $3747$ \\
\hline\hline

\multirow{6}{*}{DenseNet121} & $-$ & $0.10$ & $-$ & $-$  \\
\cline{2-5}
& {\bf HPA} & $21.50$ & $71.68$ & $50000$  \\
\cline{2-5}
{} & {\bf MPA}\_RGB & $24.90$ & $19.38$ & $28088$   \\
\cline{2-5}
{} & {\bf TPA}\_N$10$\_$2\%$ & $97.10$ & $7.87$ & $15920$ \\
\cline{2-5}
{} & {\bf TPA}\_N$10$\_$4\%$ & $99.90$ & $10.19$ & $8953$ \\
\cline{2-5}
{} & {\bf TPA}\_N$10$\_$10\%$ & $100.00$ & $15.84$ & $3970$ \\

\hline\hline

\multirow{6}{*}{ResNeXt50} & $-$ & $0.00$ & $-$ & $-$  \\
\cline{2-5}
& {\bf HPA} & $25.40$ & $72.57$ & $50000$  \\
\cline{2-5}
{} & {\bf MPA}\_RGB & $27.60$ & $13.86$ & $24738$   \\
\cline{2-5}
{} & {\bf TPA}\_N$10$\_$2\%$ & $97.60$ & $7.59$ & $15189$ \\
\cline{2-5}
{} & {\bf TPA}\_N$10$\_$4\%$ & $99.70$ & $9.60$ & $8223$ \\
\cline{2-5}
{} & {\bf TPA}\_N$10$\_$10\%$ & $100.00$ & $15.04$ & $3538$ \\

\hline\hline

\multirow{6}{*}{MobileNet-V2} & $-$ & $0.10$ & $-$ & $-$  \\
\cline{2-5}
& {\bf HPA} & $22.10$ & $69.45$ & $50000$  \\
\cline{2-5}
{} & {\bf MPA}\_RGB & $27.70$ & $16.64$ & $28294$   \\
\cline{2-5}
{} & {\bf TPA}\_N$10$\_$2\%$ & $98.50$ & $7.78$ & $15479$ \\
\cline{2-5}
{} & {\bf TPA}\_N$10$\_$4\%$ & $99.90$ & $10.39$ & $8948$ \\
\cline{2-5}
{} & {\bf TPA}\_N$10$\_$10\%$ & $100.00$ & $16.85$ & $4422$ \\

\hline

\end{tabular}}
\label{Tab:Target attack}
\end{table*}

\begin{figure}[!t]
    \centering
    \subfigure{\includegraphics[width=\textwidth]{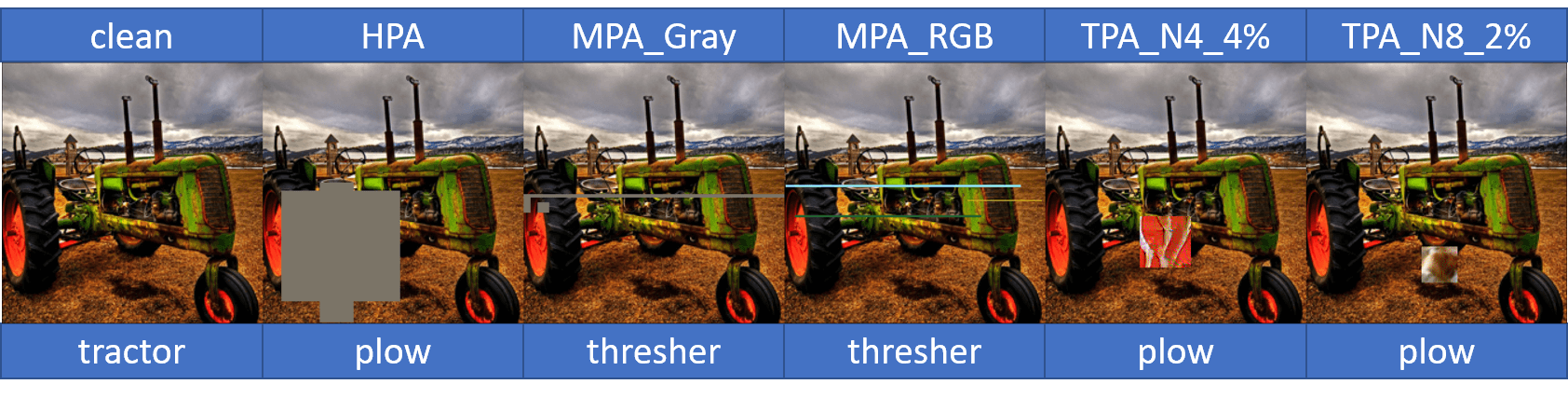}}
    \subfigure{\includegraphics[width=\textwidth]{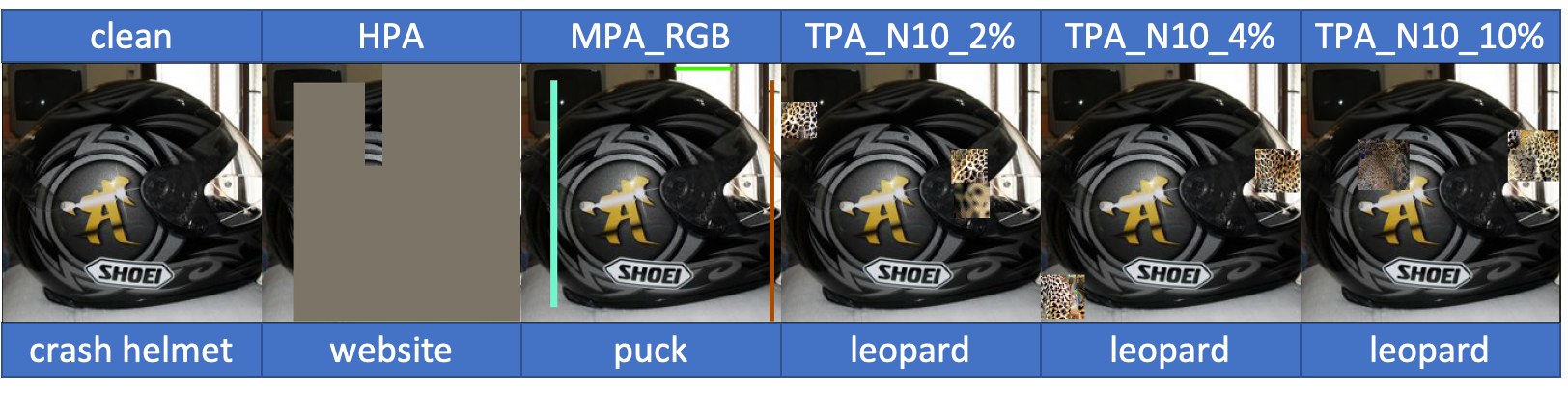}}
    \caption{Visualization of patch-attacked examples on ResNet50. The first row corresponds to non-targeted attacks and the second targeted attacks with the target class being leopard. More examples and their attention maps are provided in Appendix A and B.}
    \label{fig:non-target attack}
\end{figure}


\subsection{Texture-based Patch Attacks Against Defenses}
\label{sec:exp-defense}
This section evaluates our attacks against popular defenses.
As our MPA and TPA are new types of attacks, we first test them on traditional SOTA defense methods \cite{xie2019feature}. Another direction is to defend our attack with shape-biased network \cite{geirhos2018imagenet}, which is expected to be a good defense against our texture-based patch attack. Additionally, we perform evaluation against the Local Gradients Smoothing (LGS)~\cite{naseer2019local} specifically designed to defend against the patch-based attack in Appendix C.

\subsubsection{Defense 1: Feature Denoising.}

In this experimental part, we choose Denoise-ResNet152~\cite{xie2019feature} to perform MPA and TPA against. It is the SOTA defense against traditional perturbation-based adversarial attacks in a white-box setting. In this scenario, the attacker has access to the architecture and weights of the deep network. This is a strictly easier setting than a black-box one. PGD~\cite{madry2017towards} can only decrease the accuracy to $55.7\%$ and $45.5\%$ after $10$ and $100$ iterations, respectively. Our experimental results are summarized in Table \ref{Tab:attack against denoise network}. For non-targeted attacks, both MPA and TPA successfully attack Denoise-ResNet152. MPA reduces the accuracy from $61.6\%$ to $0.00\%$ with the occluded area only being $0.48\%$, which is even smaller than those of any our previous non-targeted attack on normal networks in Table \ref{Tab:Non-target attack}. The two versions of TPA decrease the accuracy to $1.6\%$ and $1.3\%$, respectively, both with the taken queries less than $1000$. For targeted attacks, the target accuracy for the network is $0.1\%$. Although MPA only increases this to $38.3\%$, it is higher than that of any our previous MPA attacks in targeted settings in Table \ref{Tab:Target attack}. TPA\_N$10$\_$4\%$ is able to improve the target accuracy to $94.60\%$, reflecting the vulnerability of this defense against TPA. 



\renewcommand{\colwidthA}{2.5cm}
\renewcommand{\colwidthB}{2.1cm}
\renewcommand{\colwidthC}{1.7cm}
\begin{table*}[!b]
\tiny
\caption{
Experimental results of the defenses on $1000$ images randomly selected from the ILSVRC2012 validation set. The maximum allowed query number is $10000$ and $50000$ for the non-targeted and targeted settings. Acc., T\_acc., Avg\_area, and Avg\_qry denote the classification accuracy on ground truth and target labels, average area percentage occluded by the patches, average query number, respectively}
\centering{
\small
\setlength{\tabcolsep}{0.08cm}
\begin{tabular}{|l||C{\colwidthA}|C{\colwidthB}|C{\colwidthB}|C{\colwidthC}|}
\hline
Non-target & Attack & Acc. (\%) & Avg\_area (\%) & Avg\_qry \\
\hline\hline

\multirow{4}{*}{Denoise\_ResNet152} & $-$ & $61.60$ & $-$ & $-$  \\
\cline{2-5}
{} & {\bf MPA}\_RGB & $0.00$ & $0.48$ & $9287$   \\
\cline{2-5}
{} & {\bf TPA}\_N$4$\_$4\%$ & $1.60$ & $4.71$ & $919$ \\
\cline{2-5}
{} & {\bf TPA\_N8\_10\%} & $1.30$ & $2.91$ & $867$ \\
\hline\hline

Target & Attack & T\_acc. (\%) & Avg\_area (\%) & Avg\_qry \\
\hline\hline

\multirow{5}{*}{Denoise\_ResNet152} & $-$ & $0.10$ & $-$ & $-$  \\
\cline{2-5}
{} & {\bf MPA}\_RGB & $38.30$ & $6.39$ & $27464$   \\
\cline{2-5}
{} & {\bf TPA}\_N$10$\_$2\%$ & $84.00$ & $9.73$ & $22196$ \\
\cline{2-5}
{} & {\bf TPA}\_N$10$\_$4\%$ & $94.60$ & $13.40$ & $13932$ \\
\cline{2-5}
{} & {\bf TPA}\_N$10$\_$10\%$ & $99.30$ & $20.90$ & $6920$ \\
\hline

\end{tabular}}
\label{Tab:attack against denoise network}
\end{table*}

\subsubsection{Defense 2: Against Shape-biased Network.} 
The textures of the patch play a significant role in magnifying the power of our patch attack, as shown in the comparisons between MPA and TPA in \ref{sec:exp-attack}. According to this dependence on the texture, the best defense against TPA is the model making predictions primarily based on the shapes of objects in the images instead of being largely influenced by their textures. Therefore, we consider the Shape-Network in \cite{geirhos2018imagenet} as the current best potential defense against TPA. The Shape-Network is trained on the Stylized-ImageNet, which is created by conducting style transfer on the whole training and validation sets of ImageNet, randomly changing object textures while maintaining object shapes in each image. By this design, the Shape-Network is supposed to be insensitive to textures but rely more on shapes to make inferences. Note that the construction of our texture dictionary used by TPA is also inspired by the style transfer dealing with object textures, as illustrated in Section \ref{methods_TPA}. So in principle, the Shape-Network is a very strong defense against our attacks. However, the experimental results in Table \ref{Tab:attack on the shape-biased network} show that TPAs easily confuse the Shape-Network with basically no difference as against a normal deep network. In the non-targeted setting, TPAs decrease the network's accuracy from $77.70\%$ to $0.50\%$ and $0.20\%$ with the occluded area being $5.19\%$ and $3.17\%$ for the two variants respectively. The average taken queries is $1137$. For the targeted setting, the three variants of TPAs increase the target accuracy from $0.10\%$ to $96.30\%$, $100.00\%$ and $100.00\%$, respectively. TPA\_N$10$\_$2\%$ provides the smallest occluded area $8.36\%$. TPA\_N$10$\_$10\%$ is the most query-efficient with only $3822$ taken queries but high occluded area $15.52\%$. TPA\_N$10$\_$4\%$ is the moderate choice with small occluded area $10.31\%$ and small taken queries $9229$.

\renewcommand{\colwidthA}{2.5cm}
\renewcommand{\colwidthB}{2.1cm}
\renewcommand{\colwidthC}{1.5cm}
\begin{table*}[!tp]
\tiny
\caption{
Experimental results of the defenses on $1000$ images randomly selected from the ILSVRC2012 validation set. The maximum allowed query number is $10000$ and $50000$ for the non-targeted and targeted settings. Acc., T\_acc., Avg\_area, and Avg\_qry denote the classification accuracy on ground truth and target labels, average area percentage occluded by the patches, average query number, respectively}
\centering{
\small
\setlength{\tabcolsep}{0.08cm}
\begin{tabular}{|l||C{\colwidthA}|C{\colwidthB}|C{\colwidthB}|C{\colwidthB}|}
\hline
Non-target & Attack & Acc. (\%) & Avg\_area (\%) & Avg\_qry \\
\hline\hline

\multirow{3}{*}{Shape-Network} & $-$ & $73.70$ & $-$ & $-$  \\
\cline{2-5}
{} & {\bf TPA}\_N$4$\_$4\%$ & $0.50$ & $5.19$ & $1242$ \\
\cline{2-5}
{} & {\bf TPA\_N8\_10\%} & $0.20$ & $3.17$ & $1031$ \\
\hline\hline

Target & Attack & T\_acc. (\%) & Avg\_area (\%) & Avg\_qry \\
\hline\hline

\multirow{4}{*}{Shape-Network} & $-$ & $0.10$ & $-$ & $-$  \\
\cline{2-5}
{} & {\bf TPA}\_N$10$\_$2\%$ & $96.30$ & $8.36$ & $17443$ \\
\cline{2-5}
{} & {\bf TPA}\_N$10$\_$4\%$ & $100.00$ & $10.31$ & $9229$ \\
\cline{2-5}
{} & {\bf TPA}\_N$10$\_$10\%$ & $100.00$ & $15.52$ & $3822$ \\
\hline

\end{tabular}}
\label{Tab:attack on the shape-biased network}
\end{table*}
\section{Conclusion}






In this work, we propose \textit{PatchAttack}, a powerful black-box texture-based patch attack. Our attack shows that even small textured patches are able to break deep neural networks. We model the attacking process as a reinforcement learning problem with an agent that is trained to superimpose patches onto the images in order to induce misclassification. Using monochrome patches only, we achieve a strong performance on non-targeted attack, surpassing previous work by a large margin using less queries and smaller patch areas.
After enabling the reinforcement learning agent to also use texture from an adversarial texture dictionary, PatchAttack achieves exceptional performances in both non-targeted and targeted settings. Furthermore, we show that PatchAttack breaks traditional SOTA defenses and shape-based networks.

\noindent\textbf{Acknowledgements} This work was supported in part by the Johns Hopkins University Institute for Assured Autonomy with grant IAA 80052272, National Science Foundation (NSF) grant BCS-1827427 and NSF grant CNS-18-54000.

\bibliographystyle{splncs04}
\bibliography{07_references}
\appendix
\renewcommand{\thesection}{\Alph{section}}

\section{Adversarial Examples}
\label{appendix: adv examples}
\begin{figure}[!ht]
    \centering
    \includegraphics[width=0.97\textwidth]{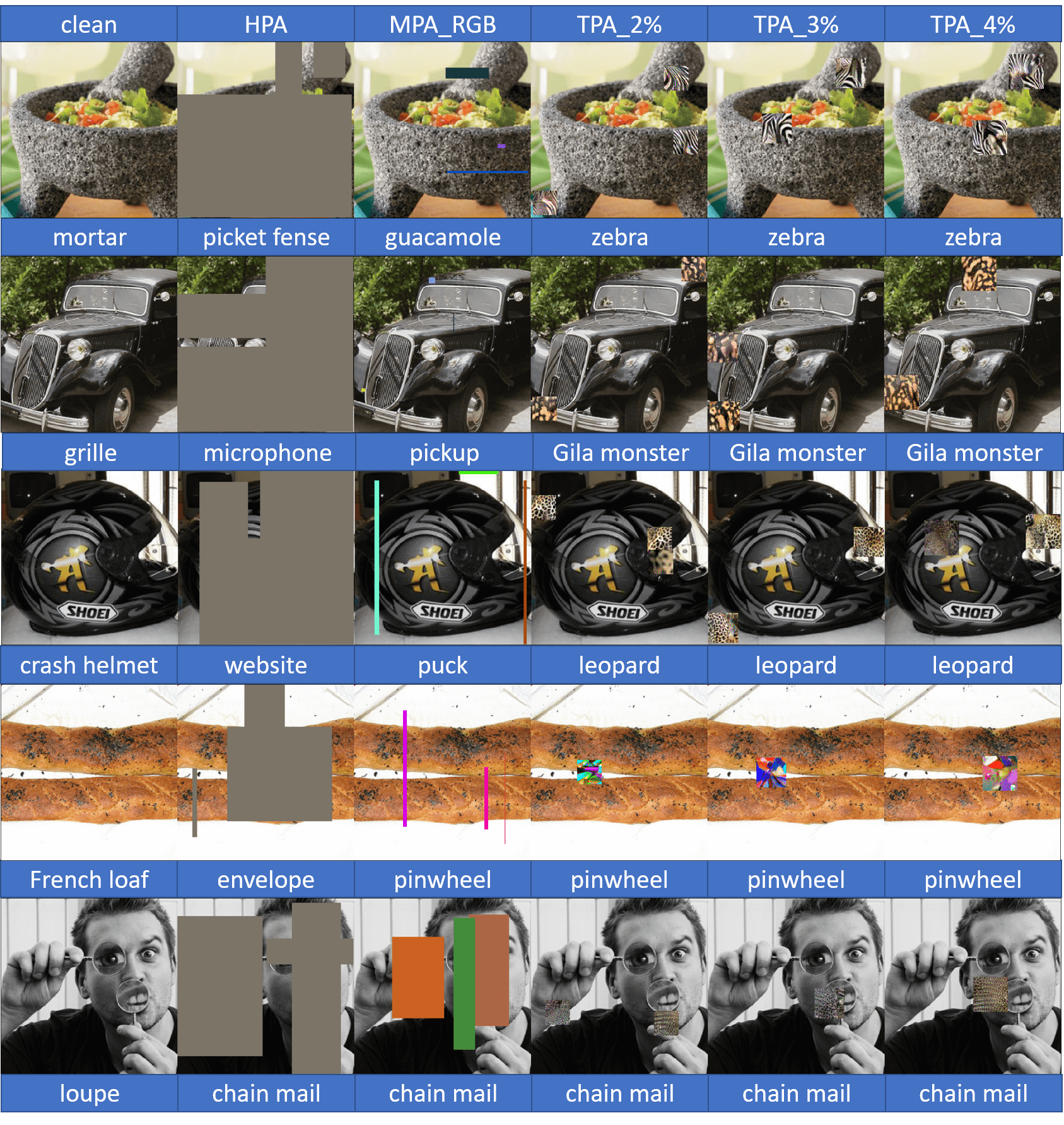}
    \caption{Adversarial examples generated by targeted \textit{PatchAttack} on ResNet50. The images in the same row are attacked with the same target class. The first three columns correspond to clean images, Hastings Patch Attack (HPA) and Monochrome Patch Attack (MPA), and the last three columns Texture-based Patch Attack (TPA) with the single patch area being $2\%$, $3\%$ and $4\%$, respectively.}
    \label{fig:adversarial examples}
\end{figure}

\newpage
\section{Attention Maps of Adversarial Examples}
\label{appendix: atn maps of adv examples}
\begin{figure}[!ht]
    \centering
    \includegraphics[width=0.97\textwidth]{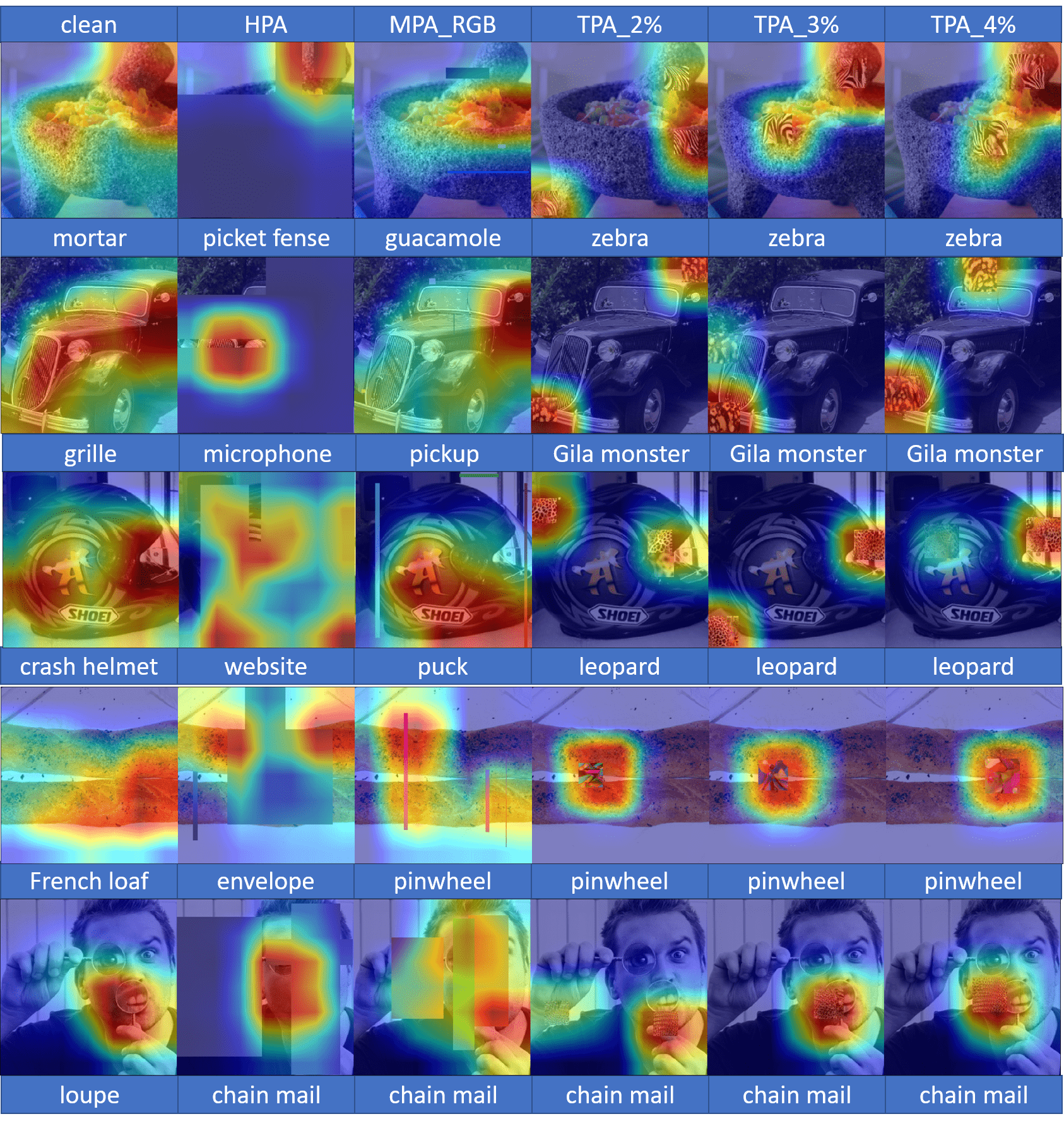}
    \caption{Attention maps of the adversarial examples in Fig.~\ref{fig:adversarial examples} generated by Grad-CAM on ResNet50. The images in the same row are attacked with the same target class. The first three columns correspond to clean images, Hastings Patch Attack (HPA) and Monochrome Patch Attack (MPA), and the last three columns Texture-based Patch Attack (TPA) with the single patch area being $2\%$, $3\%$ and $4\%$, respectively.}
    \label{fig:gradcam}
\end{figure}

\newpage

\section{Defense 3: against white-box patch attack defense}
\label{appendix: against LGS}
We evaluate our Texture-based Patch Attack (TPA) against Local Gradients Smoothing (LGS)~\cite{naseer2019local}
which is dedicated to defend against white-box patch attack on ImageNet. We perform the targeted attack on ResNet50 with the same setting in 
\ref{sec:exp-attack} 
and show the result in Table \ref{Tab:attack against LGS}. While LGS leads to slightly higher patch area and slightly lower target accuracy, it clearly fails to defend against TPA.

\renewcommand{\colwidthA}{2.5cm}
\renewcommand{\colwidthB}{2.1cm}
\renewcommand{\colwidthC}{1.5cm}
\begin{table*}[!htp]
\tiny
\caption{
Experimental results on $1000$ images randomly selected from the ILSVRC2012 validation set. T\_acc. and Avg\_area denote the classification accuracy on target labels and average area percentage occluded by the patches, respectively}
\centering{
\small
\setlength{\tabcolsep}{0.08cm}
\begin{tabular}{|C{\colwidthA}|C{\colwidthB}|C{\colwidthB}|C{\colwidthB}|}
\hline
Attack & Defense & T\_acc.(\%) & Avg\_area(\%)\\
\hline\hline
{\bf TPA}\_N$10$\_$4\%$ & -- & $99.70$ & $9.97$ \\
\cline{1-4}
{\bf TPA}\_N$10$\_$4\%$ & LGS & $97.50$ & $13.25$ \\
\hline

\end{tabular}}
\label{Tab:attack against LGS}
\end{table*}

\section{Comparison between Metropolis-Hastings sampling and Reinforcement Learning}
\label{appendix: Hastings vs RL}

We implement the Hastings Patch Attack (HPA) in the same RGB and texture search space used by Monochrome Patch Attack (MPA) and Texture-based Patch Attack (TPA)
to compare this sampling method and Reinforcement Learning method (RL). The experiments are performed on ResNet50 with the standard setup in 
\ref{sec:exp-attack}.

\renewcommand{\colwidthA}{2.5cm}
\renewcommand{\colwidthB}{2.1cm}
\renewcommand{\colwidthC}{1.5cm}

\begin{table*}[!htp]
\caption{
Experimental results of the defenses on $1000$ images randomly selected from the ILSVRC2012 validation set. The maximum allowed query number is $10000$ and $50000$ for the non-targeted and targeted settings. Acc., T\_acc., Avg\_area, and Avg\_qry denote the classification accuracy on ground truth and target labels, average area percentage occluded by the patches, average query number, respectively}
\tiny
\centering{
\small
\setlength{\tabcolsep}{0.08cm}
\begin{tabular}{|C{\colwidthA}|C{\colwidthB}|C{\colwidthB}|C{\colwidthB}|}
\hline
Non-targeted & Acc.(\%) & Avg\_area(\%) & Avg\_qry\\
\hline\hline
{\bf HPA}\_RGB & $0.20$ & $16.88$ & $10000$ \\
\cline{1-4}
{\bf MPA}\_RGB & $0.00$ & $5.41$ & $9681$ \\
\hline
\hline
targeted & T\_acc.(\%) & Avg\_area(\%) & Avg\_qry\\
\hline\hline
{\bf HPA}\_RGB & $24.80$ & $69.63$ & $50000$ \\
\cline{1-4}
{\bf MPA}\_RGB & $25.90$ & $18.45$ & $28361$ \\
\hline

\end{tabular}}

\label{Tab:H vs RL - RGB}
\end{table*}

\begin{table*}[!htp]
\caption{
Experimental results of the defenses on $1000$ images randomly selected from the ILSVRC2012 validation set. The maximum allowed query number is $10000$ and $50000$ for the non-targeted and targeted settings. Acc., T\_acc., Avg\_area, and Avg\_qry denote the classification accuracy on ground truth and target labels, average area percentage occluded by the patches, average query number, respectively}
\tiny
\centering{
\small
\setlength{\tabcolsep}{0.08cm}
\begin{tabular}{|C{\colwidthA}|C{\colwidthB}|C{\colwidthB}|C{\colwidthB}|}
\hline
Non-targeted & Acc.(\%) & Avg\_area(\%) & Avg\_qry\\
\hline\hline
{\bf HPA}\_N$4$\_$4\%$ & $1.10$ & $5.42$ & $3522.5$ \\
\cline{1-4}
{\bf TPA}\_N$4$\_$4\%$ & $0.30$ & $5.06$ & $1137$ \\
\hline
\hline
targeted & T\_acc.(\%) & Avg\_area(\%) & Avg\_qry\\
\hline\hline
{\bf HPA}\_N$10$\_$4\%$ & $99.80$ & $10.89$ & $14345$ \\
\cline{1-4}
{\bf TPA}\_N$10$\_$4\%$ & $99.70$ & $9.97$ & $	8643$ \\
\hline
\end{tabular}}

\label{Tab:H vs RL - Texture}
\end{table*}

It is observed that MPA\_RGB is better than HPA\_RGB, because it achieves lower accuracy in the non-targeted setting and higher target accuracy in targeted setting, while also using a smaller area and less queries.

Here we can observe that RL still is much more query-efficient than the sampling algorithm, however, the methods are comparable in terms of accuracy and occlusion area. This can be attributed to our improved search space for performing the attacks, highlighting the importance of our texture dictionary.

\section{Transferability of adversarial patch dictionary generated by white-box method}
\label{sec:transfer}
We implement Adversarial Patch (AP)~\cite{brown2017adversarial}, 
the white-box patch attack. We first generate an adversarial patch dictionary (AdvPatchDict) consisting of $1000$ classes by attacking VGG19 using AP on ImageNet dataset, and then attack the other $4$ networks used in our experiments with those patches in the dictionary. The results are shown in the Table \ref{Tab:AdvPatchDict}. In non-targeted settting, AdvPatchDict decreases accuracy to $0.20\%$ on VGG19 but only to $56\% - 66\%$ on the other networks. In targeted setting, it increases target accuracy on VGG to $98.20\%$ but basically fails to increase it for other networks. Clearly, AdvPatchDict generated by the white-box method overfits to the architecture used to generate them, highlighting the superiority of the design of our texture dictionary.

\renewcommand{\colwidthA}{2.5cm}
\renewcommand{\colwidthB}{2.5cm}
\renewcommand{\colwidthC}{2.0cm}
\begin{table*}[!htp]
\tiny
\caption{
Experimental results on $1000$ images randomly selected from the ILSVRC2012 validation set. Acc., T\_acc. and P\_area, denote the classification accuracy on ground truth and target labels, area percentage occluded by the adversarial patch, respectively}
\centering{
\small
\setlength{\tabcolsep}{0.08cm}
\begin{tabular}{|C{\colwidthA}|C{\colwidthB}|C{\colwidthB}|C{\colwidthC}|}
\hline
AdvPatchDict & Non-targeted Acc.(\%) & targeted T\_acc.(\%) & P\_area(\%)\\
\hline\hline
VGG19 & $0.20$ & $98.20$ & $8.95$ \\
\cline{1-4}
ResNet50 & $62.50$ & $0.00$ & $8.95$ \\
\cline{1-4}
DenseNet121 & $57.80$ & $1.70$ & $8.95$ \\
\cline{1-4}
ResNeXt50 & $65.30$ & $0.10$ & $8.95$ \\
\cline{1-4}
MobileNet-V2 & $56.00$ & $0.10$ & $8.95$ \\
\hline

\end{tabular}}
\label{Tab:AdvPatchDict}
\end{table*}

\end{document}